\documentclass[journal]{IEEEtran}
\ifCLASSINFOpdf
\else
\fi
\usepackage{kotex}
\usepackage{kotex-logo}
\usepackage{multirow}
\usepackage{array}
\usepackage{makecell}
\usepackage{graphicx} 
\usepackage{subfig}

\begin{document}
%
\title{Color Information-Based Automated Mask Generation for Detecting Underwater Atypical Glare Areas}
%
%
%

\author{Mingyu~Jeon$^{1}$,
        Yeonji~Paeng$^{1}$,
        and~Sejin~Lee$^{2}$

\thanks{$^{1}$ Mingyu Jeon and Yeonji Paeng are with the Department of Mechanical Engineering, Kongju National University, 1223-24 Cheonan-daero, Cheonan 31080, Republic of Korea. {\tt\small \{\texttt{jmk208435, pang9434}@smail.kongju.ac.kr\}}}

\thanks{$^{2}$ Sejin Lee is with the Division of Mechanical and Automotive Engineering, Kongju National University, 1223-24 Cheonan-daero, Cheonan 31080, Republic of Korea. {\tt\small sejiny3@kongju.ac.kr}}%

\thanks{This research was supported by Development of standard manufacturing technology for marine leisure vessels and safety support robots for underwater leisure activities of Korea institute of Marine Science \& Technology Promotion (KIMST) funded by the Ministry of Oceans and Fisheries(KIMST-20220567).}

\thanks{Manuscript uploaded February 22, 2025.}}

\maketitle
\begin{abstract}
Underwater diving assistance and safety support robots acquire real-time diver information through onboard underwater cameras. This study introduces a breath bubble detection algorithm that utilizes unsupervised K-means clustering, thereby addressing the high accuracy demands of deep learning models as well as the challenges associated with constructing supervised datasets. The proposed method fuses color data and relative spatial coordinates from underwater images, employs CLAHE to mitigate noise, and subsequently performs pixel clustering to isolate reflective regions. Experimental results demonstrate that the algorithm can effectively detect regions corresponding to breath bubbles in underwater images, and that the combined use of RGB, LAB, and HSV color spaces significantly enhances detection accuracy. Overall, this research establishes a foundation for monitoring diver conditions and identifying potential equipment malfunctions in underwater environments.
\end{abstract}


%
\IEEEpeerreviewmaketitle

\section{Introduction}
%
%
%
%
Underwater diving activity assistance and safety support robots perform diver information collection and diver tracking using underwater sensors and thrusters.
In this process, wearable devices attached to the diver's body or equipment can be used for diver information acquisition, while underwater cameras and sonar sensors can be utilized for long-distance data collection.
Among these, using cameras generally enables real-time monitoring, provides intuitive output images, and allows for the identification of fine details such as the texture of nearby objects.
This enables the collection of information such as the diver's posture, the condition of the diving equipment, and bubbles generated by the diver's breathing.
By analyzing this information, it is possible to infer the diver’s hand signals, abnormal behavior, and any malfunctions in the diving equipment.
In particular, analyzing breath bubble information allows for the estimation of the diver’s breathing cycle and volume, which can further be used as a factor in assessing the diver's diving capability.
At the same time, the bubble information included in the image provides insights into the diver’s condition, which is the primary observation target.
Therefore, an algorithm capable of detecting bubbles within image frames is essential for utilizing image and video data collected via cameras for diver monitoring.
Deep learning model-based algorithms can provide high accuracy; however, they require a large dataset for supervised learning.
Constructing such a dataset requires significant time and resources.
In particular, bubbles in underwater images generally exhibit dynamic and irregular shapes and boundaries, significantly increasing the difficulty of constructing training datasets.
This study proposes a breath bubble detection method utilizing K-means clustering, an unsupervised machine learning algorithm, by enhancing clustering-based results through image relative coordinate information and image color space fusion.

\section{Related Works}
\subsection{Underwater Camera Images}
Underwater environments and objects captured using underwater cameras possess distinct characteristics that differentiate them from images taken in the atmosphere with conventional cameras.
The primary difference between the two imaging environments lies in the medium; underwater cameras operate in water, where strong refraction and reflection occur more easily than in air.
This difference causes the rapid absorption of long-wavelength red electromagnetic waves compared to blue waves, resulting in underwater RGB images where the Red channel values are relatively lower than those of the Green and Blue channels. As the distance increases, the number of pixels lacking Red channel values also increases.
This leads to an imbalance in the information distribution among the three channels in an RGB image.
Furthermore, in underwater environments, camera output is affected by backscattering, a phenomenon where light is scattered by suspended particles or fine debris.
\begin{figure*}[ht]
    \centering
    \includegraphics[width=\textwidth]{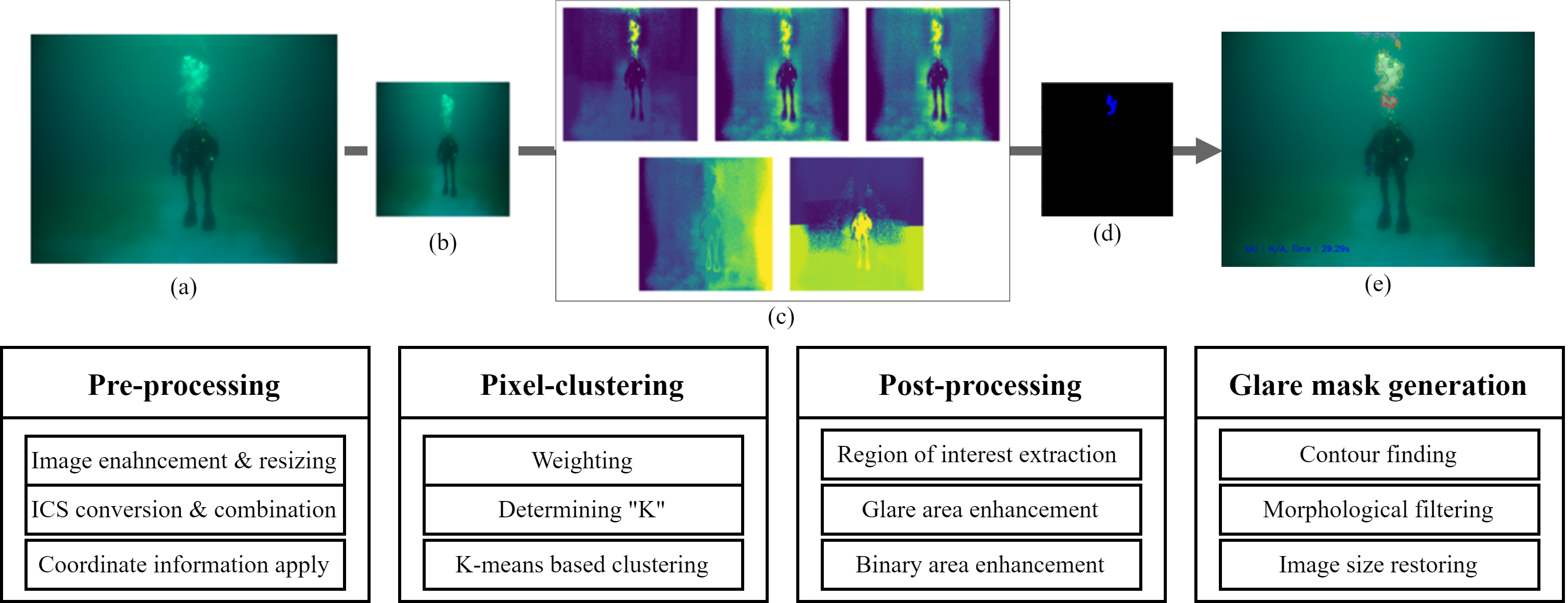}
    \caption{Flow of the proposed glare detection algorithm and input/output examples of each phase: (a) Input image; (b) Image after pre-processing (brightness enhancement, image color space combination, coordinate information channel supplementation, and image resizing); (c) Pixel-clustering results showing clusters per channel; (d) Binary region obtained using cluster information; (e) Output image.}
    \label{fig}
\end{figure*}
Such backscattering degrades the quality of captured images, making them appear blurry and hazy.
Additionally, underwater environments contain various suspended particles, turbidity, and bubbles, which further impact image quality.
These factors introduce noise into the captured images, negatively affecting image analysis.
Due to these environmental characteristics, underwater camera images require preprocessing and correction techniques to mitigate these effects.
Existing studies that utilize underwater camera data primarily focus on detecting divers, natural or artificial structures in underwater environments, and further detecting diver postures and hand signals.
Therefore, glare is typically treated as simple noise or an obstacle that negatively affects detection, rather than being considered a detection target in most research efforts.

\subsection{Data Clustering}
Data clustering is a widely used technique not only in data processing fields such as feature analysis and data mining but also in most formats containing information, such as images.
Representative clustering methods include K-means, Mean Shift, and Gaussian Mixture Model (GMM). Among these, the K-means clustering algorithm first designates K clusters in the dataset before performing clustering and assigns centroids corresponding to these clusters.
These cluster centroids calculate the distance to each individual data point, and each data point is assigned to the centroid with the minimum distance.
The newly formed cluster’s mean point is then set as the new centroid, and the clustering assignment process is repeated.
When the updated centroids stabilize within a certain threshold, the clustering process terminates, and the final results are produced.
When directly applied to images, this type of clustering groups individual data points based on pixel color values.
However, in such pixel clustering, only color values are considered, and the two-dimensional spatial information of each pixel within the image is not incorporated.
As a result, adjacent pixels within the same object may be misclassified into different clusters due to improper clustering results, or distant pixels from unrelated objects that share similar color values may be grouped into the same cluster.

\subsection{CLAHE}
CLAHE (Contrast Limited Adaptive Histogram Equalization) is a type of histogram equalization used to enhance image contrast.
HE (Histogram Equalization) adjusts contrast across the entire image. In contrast, CLAHE divides the image into a grid structure and applies HE individually to each grid.
At this stage, contrast enhancement is limited to prevent it from exceeding a certain threshold.
Through this process, while HE can degrade image quality in areas with locally high contrast, CLAHE enhances the quality of low-contrast areas by improving unclear shapes and boundaries while suppressing excessive contrast enhancement, thus improving overall image quality.
CLAHE improves image quality in underwater camera images by addressing issues such as backscattering and insufficient light, which cause unclear shapes and boundaries. At the same time, it suppresses excessive local contrast caused by scattering and reflection.

\subsection{Image Color Spaces}
Image color space refers to a method of representing color images, which typically consists of three two-dimensional channels organized in rows and columns.
The most commonly used image color space is RGB (Red, Green, Blue), which has a color intensity range from 0 to 255 and is easily processed by hardware.
The HSV (Hue, Saturation, Value) color space represents hue values ranging from 0 to 360, while saturation and value range from 0 to 100.
Additionally, various color spaces exist, such as the Lab color space, which consists of luminance, green-red range, and blue-yellow range, and the YUV color space, which consists of brightness, cyan-blue, and red-yellow levels.
By using these image color spaces, image adjustment can be performed more easily by assigning different weights to each channel. Additionally, separating channels allows for the extraction of meaningful information and features from each channel.

This study draws its core concept from the simple yet powerful clustering capability of pixel clustering, which applies data clustering techniques to image pixels.
It is proposed that the issue of losing spatial relationships between pixels during the pixel clustering process can be mitigated by adding additional channels representing image coordinates.
Furthermore, noise introduced into images due to the underwater environment is addressed using computer vision techniques such as CLAHE.

\section{Algorithm Structure}
The glare detection algorithm for breath bubbles can be seen in Fig. 1. The algorithm is structured into three main stages.
The first stage is preprocessing, where image resizing, image color space fusion, CLAHE application, and image channel expansion are performed.
The second stage is pixel clustering, in which K-means-based pixel clustering is applied to the five-channel image processed in the previous stage, creating a five-channel image composed of K clusters.
The final stage is the post-processing step, where the glare regions consistently identified in both the color channel and relative coordinate channel are extracted.

\subsection{Preprocessing}
The preprocessing stage primarily involves image color space transformation and fusion, image resizing, image correction, and the addition of relative coordinate information.
The RGB color space of images captured in underwater environments contains very little information in the Red channel, making it less useful for meaningful data extraction.
Such insignificant channels are replaced with meaningful information, such as the Value channel in the HSV color space and the Luminance channel in the Lab color space, to address the imbalance in information across different channels.
The K-means clustering algorithm repeatedly computes the distance between each data point and the K cluster centroids.
As a result, the higher the image resolution, the greater the computational load required.
High-resolution images typically provide better quality, but they can lead to excessive segmentation in the pixel clustering process.
To prevent such issues and achieve an optimal computational time, image resizing is applied to reduce the resolution.
Subsequently, CLAHE is applied to prevent excessive contrast enhancement in glare regions while improving the unclear boundaries between breath bubble regions and the background.

\begin{figure}[h]
    \centering
    \includegraphics[width=0.5\textwidth]{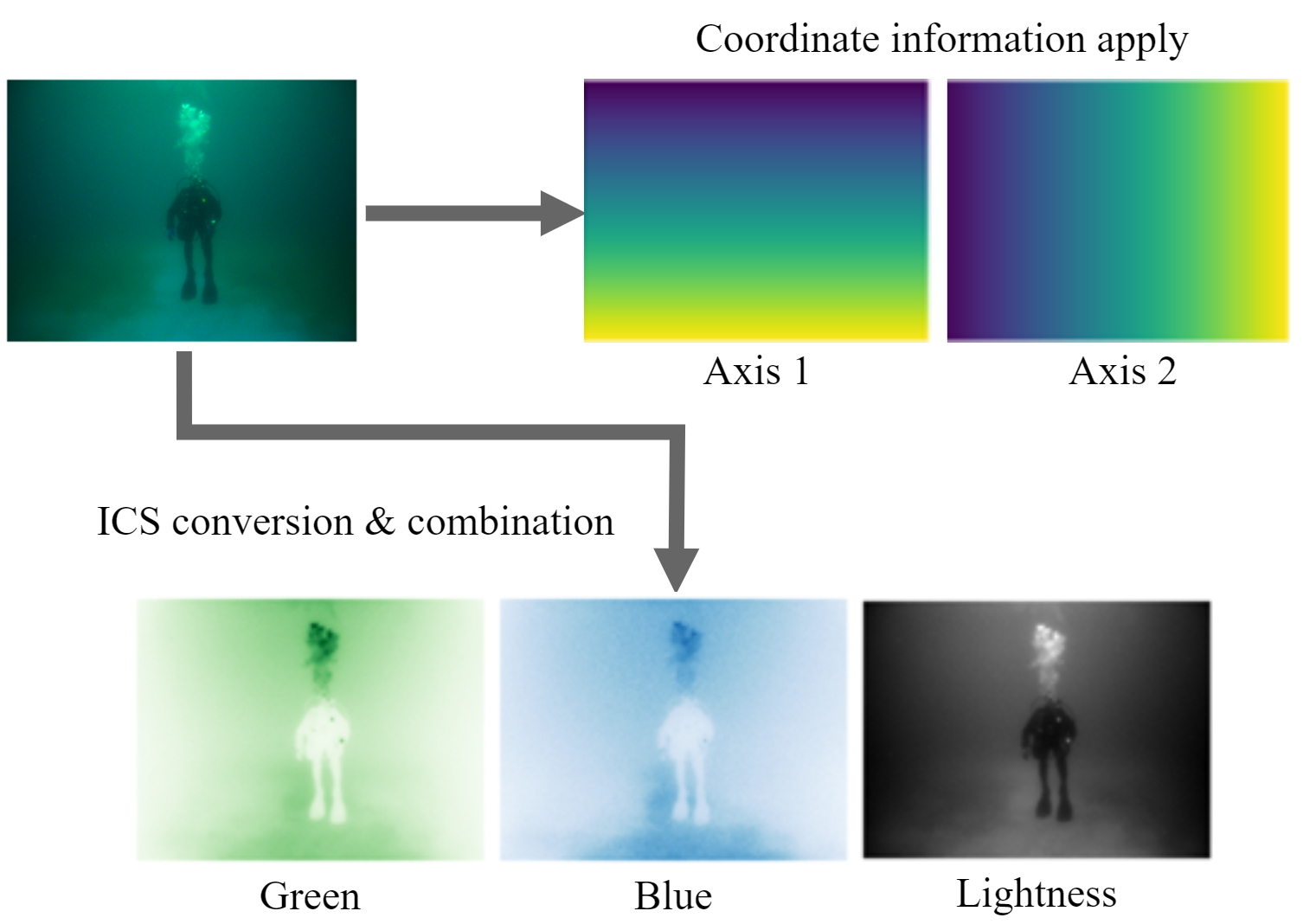}
    \caption{The original RGB image input into the preprocessing phase and the phase output, which includes the GBL image converted to a color space with three channels, along with the X and Y coordinate information channels.}
    \label{fig}
\end{figure}

Finally, two additional channels containing row and column coordinate information are added to provide distance information between pixels during the pixel clustering process, resulting in a five-channel image structured as shown in Fig. 2.

\subsection{Pixel Clustering}
The pixel clustering stage applies clustering to the five-channel image processed in the preprocessing stage. This stage mainly consists of assigning weights to each channel and performing the K-means algorithm.
The transformed and fused image color space from the preprocessing stage may exhibit differing value ranges and characteristics depending on its intended use.
\begin{figure}[h]
    \centering
    \includegraphics[width=0.5\textwidth]{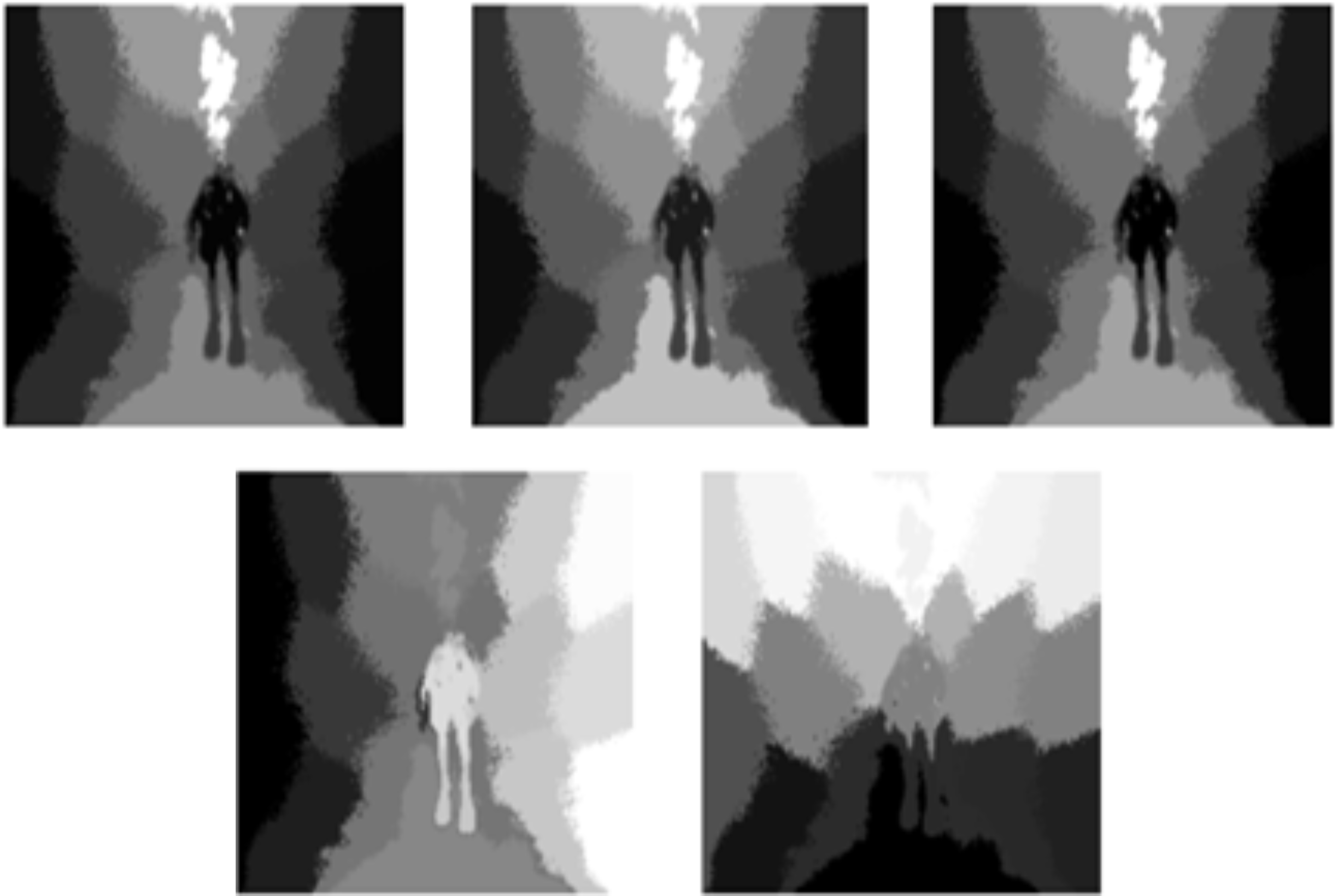}
    \caption{Clustered images for each channel generated by the pixel clustering phase. }
    \label{fig}
\end{figure}
To designate clusters for specific regions, the value range for each channel is defined to align with consistently observed features in those regions.
The K-means clustering algorithm in the pixel clustering stage follows the steps outlined below.
First, the five-channel image is converted into a one-column data format.
At this point, the data is restructured from its original row × column format to a 1 × (row × column) format. The fourth and fifth channels, which contain relative spatial information, are transformed into a single-row format to incorporate pixel location data in the clustering process.
Next, to determine whether glare is present in the image, the number of clusters K is determined using the following equation.

\begin{equation}
K = \left\lfloor \frac{(V_{\max} - V_{\min})^2}{266} \right\rfloor
\end{equation}

This approach is based on the characteristic that glare in underwater images generally has a high Value, while the underwater background has a low Value.
K data points are randomly selected as initial centroids.
According to the equation below, the Euclidean distance between each data point and the K centroids is computed, and each data point is assigned to the cluster associated with its nearest centroid.

\begin{equation}
\arg\min \sum_{i=1}^{K} \sum_{x_j \in C_i} \left\| x_j - c_i \right\|^2
\end{equation}

Next, the mean of all data points assigned to each cluster is calculated, and this mean is set as the new cluster centroid.
Then, the Euclidean distance between the newly computed centroid and the previous centroid is calculated, and the cluster centroid is updated accordingly.
This process is repeated until the centroids converge.
If the distance between centroids falls below a predefined threshold, the centroid is finalized as the clustering result.

As shown in Fig. 3, the result of the pixel clustering stage consists of five channels identical to the input, with each channel composed of K clusters.
The boundaries between clusters are identical across all channels, but the representative value of each cluster varies depending on the image data and assigned weights of each channel.
Thus, by analyzing the clustering results for each channel separately, the tendencies of each cluster within the respective channels can be identified.
For example, in the three channels containing color information, the representative value of each cluster corresponds to the distribution of colors, while in the two channels containing row and column coordinate information, the representative value is determined based on the relative spatial orientation of each cluster.

\subsection{Post-processing}
In the final stage, the post-processing phase extracts glare regions by performing difference operations between the color channels and the relative coordinate channels.
A high cluster representative value in the color channel corresponds to clusters that include regions where reflection and scattering occur, such as breath bubbles and the bottom surface below the water.
To separate the high-contrast regions of the breath bubble area from the high-contrast regions of the bottom surface or other objects, a difference operation is performed between the results of the relative coordinate channel and the color channel.
\begin{figure}[h]
    \centering
    \includegraphics[width=0.5\textwidth]{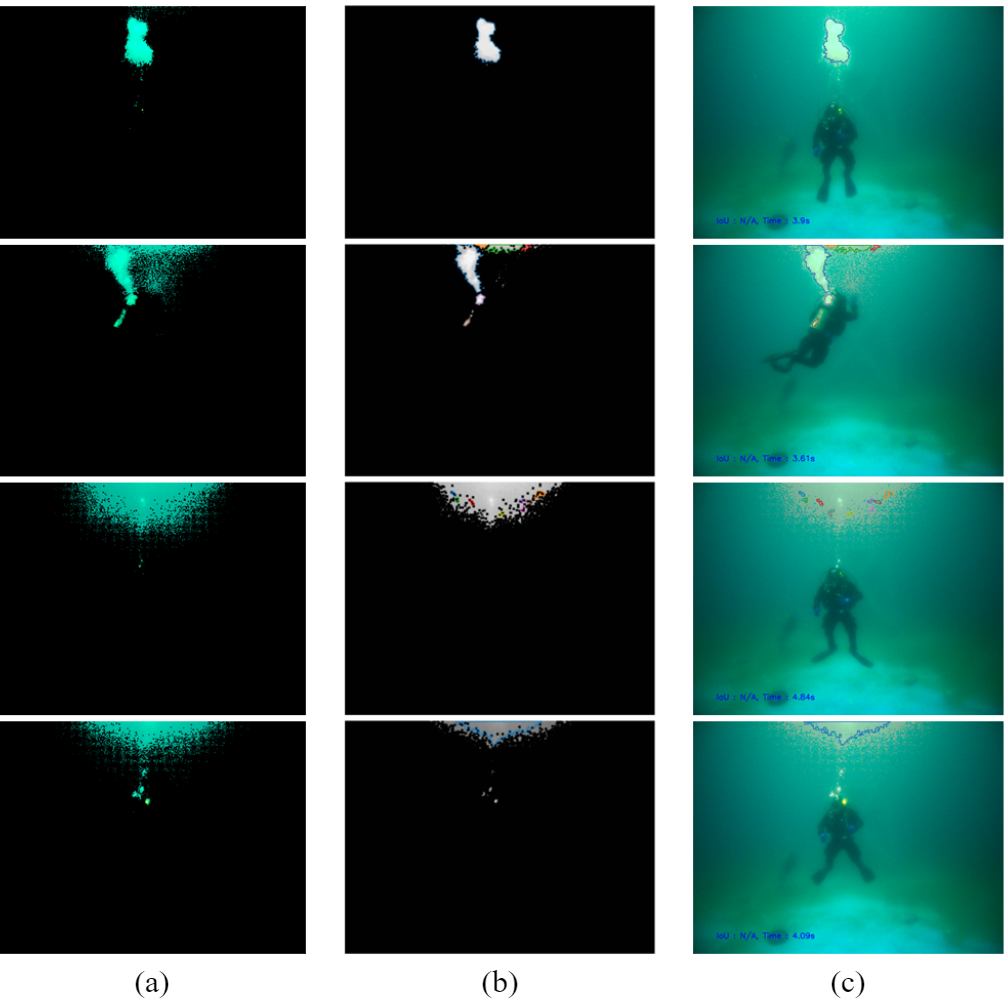}
    \caption{ Example of the sub-mask generation process in the post-processing phase: (a) ensembled result, (b) result after applying erosion-based filtering, instance assignment using the marching squares technique, and area-based filtering, (c) result overlaid on the original input image.}
    \label{fig}
\end{figure}
A binary mask result, as shown in Fig. 4(a), is obtained by computing the non-overlapping regions between the cluster areas in the lower region where the diver is not present in the relative coordinate channel and the high-contrast cluster areas in the color channel.
The breath bubble region obtained through this process has an improved boundary between the object-free background and the glare area through CLAHE. However, the boundary between relatively bright areas caused by surface light sources and the breath bubble region exhibits relatively low contrast.
Thus, significant noise and false detections occur during the rising and expansion stages of the breath bubbles.
To improve these results, the following sub-mask generation steps are applied.
\begin{figure}[h]
    \centering
    \includegraphics[width=0.5\textwidth]{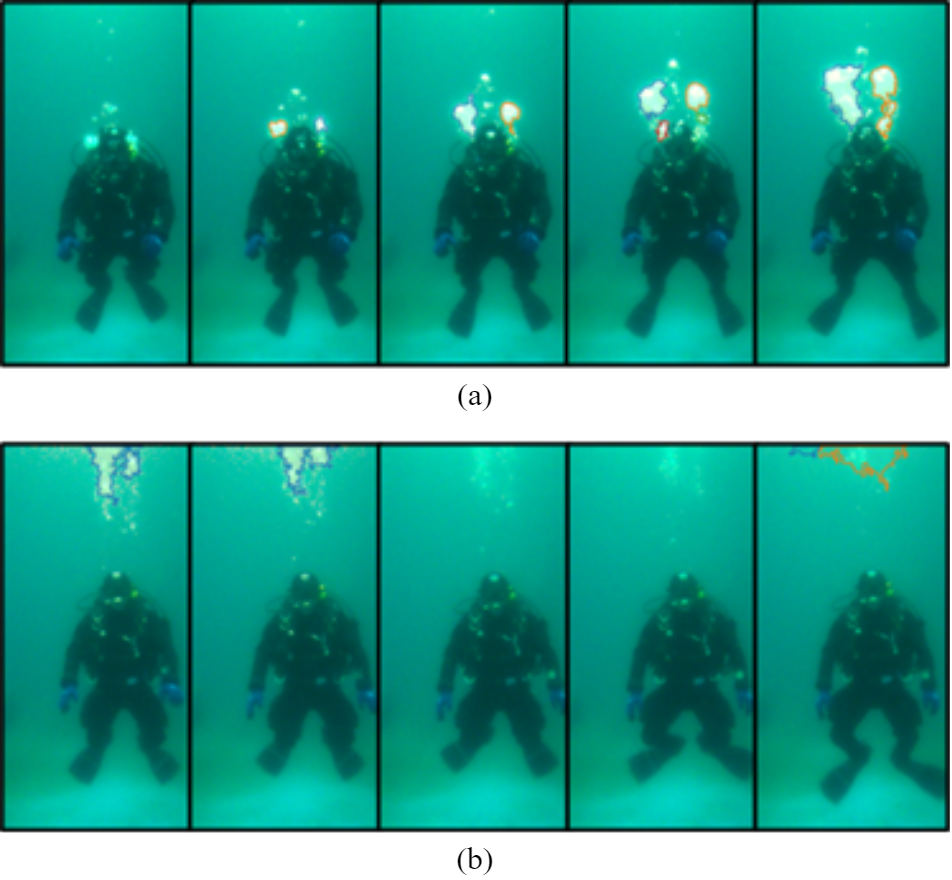}
    \caption{ The images of the respiration bubble formation, expansion, and dissolution stages: 10 consecutive frames with results shown at 2-frame intervals. (a) Results of the respiration bubble formation and expansion stages, (b) Results of the respiration bubble dissolution stage.}
    \label{fig}
\end{figure}
The sub-mask is used before the final glare is returned to filter out excessively sensitive detections, such as small pixel clusters from floating debris and background, as well as excessive detection regions in the water surface during the bubble dissipation phase.
First, the boundaries of the binary mask are refined using an erosion technique to reduce background regions not eliminated at the bubble boundaries and to minimize small noise clusters resulting from the clustering process.
Second, the binary mask is intersected with the original image to utilize fine details present in the color channels of the original image.
In the final step, the extracted region is converted into a grayscale image, and the region averaging technique is applied to output the improved region as the white area in Fig. 4(b).
The sub-mask generation based on contour extraction using the Marching Squares algorithm follows the steps outlined below.

 1)	The given grey-scale image is divided into a 2×2 grid, assigning image pixel values to each vertex.
 
 2)	For each grid, binary results are assigned to each vertex based on a global threshold, determined by the average value of the entire image.
 
 3)	Based on the combination of binary values from the four vertices, the contour and region creation results for each grid are assigned.
 
 4)	Using the results from each grid, continuous contour lines are generated by determining the connectivity of line segments.
 
 5)	The number of pixels in each contour is designated as its area.

Particularly, during the process of assigning separate contours to distinct regions, an instance is assigned to each region.
An allowable range of area is specified for each instance to apply noise filtering to over-clustered areas where background and breath bubbles are mixed, as well as small background pixels misclassified due to underwater floating particles.
The results after this filtering process are represented as contour shapes in Fig. 4(b), while areas without contour representation indicate suppression by area-based filtering.
Through this process, only the acquired contrast areas are output as the final regions, as shown in Fig. 4(c).

\section{Experiments and Results Analysis}
To qualitatively analyze the results of the proposed breath bubble glare detection algorithm, experiments were conducted using experimental data obtained in an indoor water tank environment.
In the indoor water tank environment, an underwater diver and the underwater camera capturing the diver were positioned at distances ranging from a minimum of 1 meter to a maximum of 5 meters, during which free swimming and hand signaling were performed.

\subsection{Qualitative Evaluation}
The qualitative evaluation of the proposed breath bubble glare detection algorithm was conducted on a scenario consisting of 2,165 consecutive frames recorded in the indoor water tank environment.
As the diver naturally breathes in the underwater environment, the breath bubbles appear repeatedly in the stages of formation, expansion, and dissipation.
Fig. 5 visualizes part of the algorithm's results on breath bubbles in the formation and expansion stages at 2-frame intervals.
It was confirmed that glare was appropriately detected in consecutive frames of breath bubbles in the formation and expansion stages.
However, in some images during the dissipation stage, the high-contrast regions formed by the merging of surface light sources and dissipating bubbles caused over-detection and under-detection.

\subsubsection{Effect of Relative Coordinate Information}
To qualitatively assess the effect of incorporating relative coordinate information, the proposed algorithm adjusted the weights of the X and Y channels in the pixel clustering stage and excluded these channels from the K-means algorithm.
Fig. 6 presents the corresponding results.
The third column represents results where relative coordinate information was correctly applied, leading to the classification of non-adjacent pixels with similar colors into different clusters.

\begin{figure}[h]
    \centering
    \includegraphics[width=0.5\textwidth]{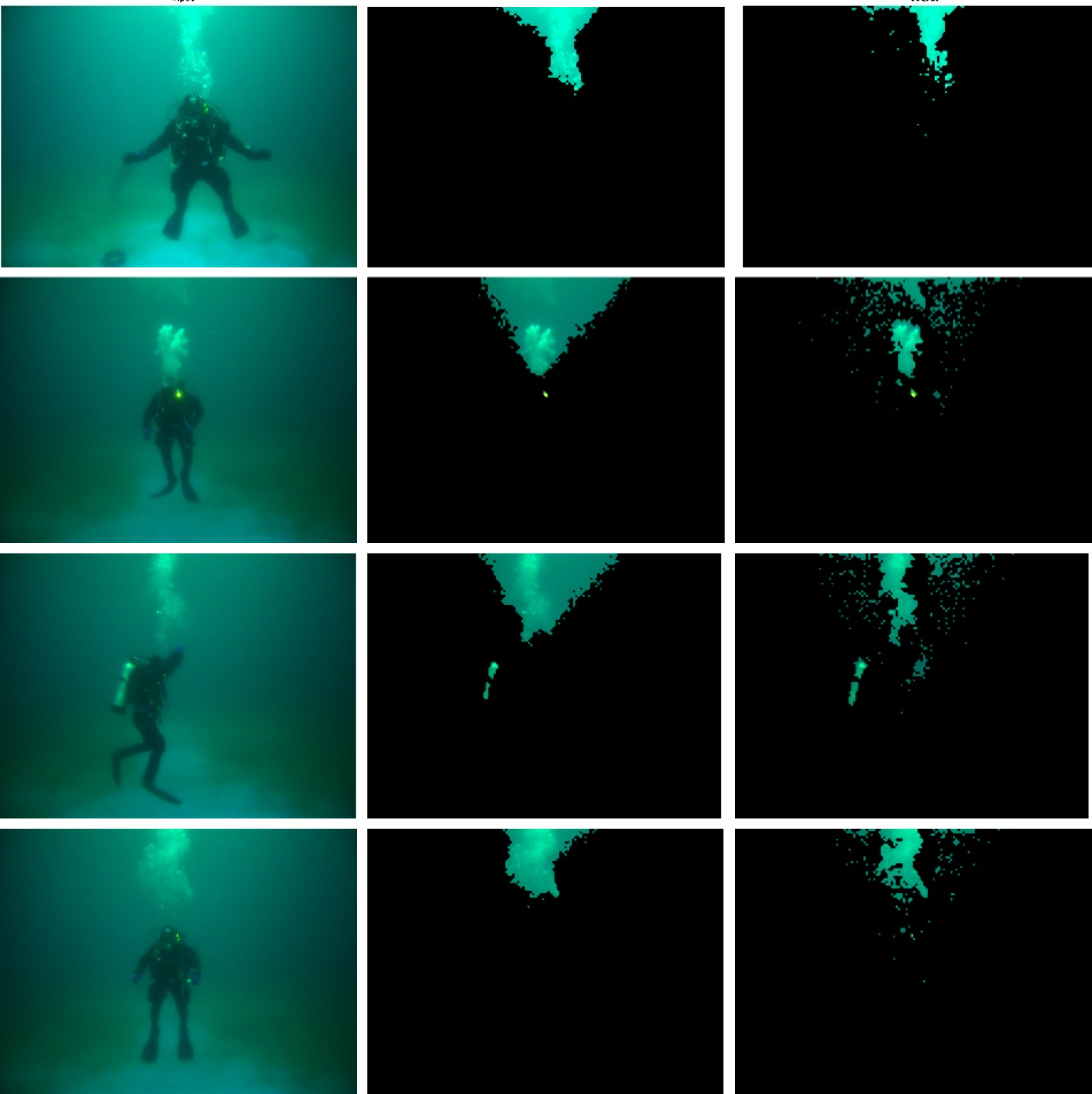}
    \caption{From left to right: original image, result without 
XY coordinate information, result with XY 
coordinate information included. }
    \label{fig}
\end{figure}

\begin{figure}[h]
    \centering
    \includegraphics[width=0.5\textwidth]{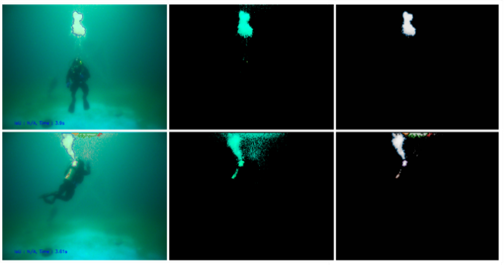}
    \caption{From left to right: original image, result without 
erosion-based filtering, result with erosion
based filtering applied. }
    \label{fig}
\end{figure}
However, in the second column, where relative coordinate information was excluded, spatially separated pixels with similar colors were assigned to the same cluster, and it was not possible to define exclusion zones for glare detection, resulting in floor reflections being included in the detection output.

\subsubsection{Effect of Erosion-Based Filtering Application}
To qualitatively assess the impact of applying erosion-based filtering, this process was omitted during the post-processing stage.
Fig. 7 presents the results comparing images with and without the application of erosion-based filtering.
This filtering process showed that, in the clustering stage, pixels corresponding to glare and background with similar colors, despite being relatively close in distance, were assigned to the same cluster.
In contrast, after applying the filtering, noise dispersed throughout the background was effectively removed, and more refined cluster boundaries were obtained.

\subsection{Qualitative Evaluation}
To quantitatively analyze the performance of the proposed algorithm, labeling was performed on 312 images.
Labeling was carried out by reducing the brightness to simplify ambiguous glare boundaries and manually defining polygon-type masks for visually identifiable boundaries.
To quantify the detection results, the detected areas from the algorithm and the ground truth areas were converted into IoU (Intersection over Union) scores for calculation.

\begin{table}[]
\caption{The mean IoU score results based on the combination of image color spaces and the application of sub-mask generation}
\centering
\begin{tabular}{c|cc}
\hline
\multicolumn{1}{c|}{\makecell{\\Image color space}}
 & \multicolumn{2}{c}{mIoU score}                                                                                                        \\ \cline{2-3} 
                                            & \multicolumn{1}{c|}{\begin{tabular}[c]{@{}c@{}}w/o sub-mask\end{tabular}} & \begin{tabular}[c]{@{}c@{}}w/ sub-mask\end{tabular} \\ \hline
RGB                                         & \multicolumn{1}{c|}{0.4276}                                                 & 0.6204                                                  \\ \hline
HSV                                         & \multicolumn{1}{c|}{0}                                                      & 0                                                       \\ \hline
Lab                                         & \multicolumn{1}{c|}{0.4699}                                                 & 0.4007                                                  \\ \hline
YUV                                         & \multicolumn{1}{c|}{0.0187}                                                 & 0                                                       \\ \hline
(R)GB+(HS)V                                 & \multicolumn{1}{c|}{0.4849}                                                 & 0.6279                                                  \\ \hline
(R)GB+L(ab)                                 & \multicolumn{1}{c|}{0.4859}                                                 & \textbf{0.6562}                                         \\ \hline
(R)GB+L(ab)+(HS)V                           & \multicolumn{1}{c|}{\textbf{0.5490}}                                        & 0.6032                                                  \\ \hline
(R)B(G)+L(ab)+(HS)V                         & \multicolumn{1}{c|}{0.4989}                                                 & 0.6522                                                  \\ \hline
\end{tabular}
\end{table}

\begin{table}[]
\caption{The result in the undetection rate via different 
combinations of image color spaces combination}
\centering
\begin{tabular}{c|cc}
\hline
\multicolumn{1}{c|}{\makecell{\\Image color space}} 
 & \multicolumn{2}{c}{Undetection rate {[}\%{]}}\\ \cline{2-3} 
                                            & \multicolumn{1}{c|}{IoU \textless 0.1} & IoU \textless 0.4 \\ \hline
RGB                                         & \multicolumn{1}{c|}{16.67}             & 22.76             \\ \hline
HSV                                         & \multicolumn{1}{c|}{100}               & 100               \\ \hline
Lab                                         & \multicolumn{1}{c|}{33.3}              & 50.0              \\ \hline
YUV                                         & \multicolumn{1}{c|}{100}               & 100               \\ \hline
(R)GB+(HS)V                                 & \multicolumn{1}{c|}{14.42}             & 21.15             \\ \hline
(R)GB+L(ab)                                 & \multicolumn{1}{c|}{10.26}             & 18.91             \\ \hline
(R)GB+L(ab)+(HS)V                           & \multicolumn{1}{c|}{19.23}             & 24.04             \\ \hline
(R)B(G)+L(ab)+(HS)V                         & \multicolumn{1}{c|}{\textbf{8.23}}     & \textbf{18.27}    \\ \hline
\end{tabular}
\end{table}

The intersection represents the overlapping area between the two regions, while the union refers to their combined area.
IoU score is calculated according to the following equation.

\begin{equation}
\textit{IoU Score} = \frac{\textit{Intersection}}{\textit{Union}}
\end{equation}

IoU scores range between 0 and 1, where values closer to 0 indicate that there is no overlap between the two regions, while a score of 1 signifies perfect alignment.
The IoU score for each image and the mean IoU (mIoU) score, which represents the average IoU across all images in the dataset, are computed using the following equation.

\begin{equation}
\textit{mIoU score} = \frac{1}{N} \sum_{i=1}^{N} IoU_{i}
\end{equation}

N represents the number of images used for evaluation.
In addition, 0.1 and 0.4 thresholds were established for IoU scores. Any result below these thresholds was considered undetected for that image. 
The undetection rate in Table.2 is calculated using the following equation. IoU$_i$ represents the IoU score of the i-th image.

\begin{equation}
\textit{Undetection Rate} = \frac{1}{N} {\sum\limits_{i=1}^{N} 1 (IoU_i < threshold)} \times 100 \, [\%]
\end{equation}

The combination of the Blue and Green channels from the RGB color space with the Lightness channel from the Lab color space achieved the highest mean IoU score of 0.6562 across the 312 images. A similar performance was observed in the BLV combination (Blue, Lightness, and HSV Value channels), which recorded an mIoU of 0.6522.
Under the same conditions except for the image color space combinations, an experiment was conducted to compare the mIoU scores for each image color space combination to analyze the impact of color space combinations.
It was observed that when using color spaces such as HSV or YUV, where each channel conveys entirely different types of information, the detection performance was inadequate due to the lack of consistent glare information.
In contrast, when combining image color space channels with similar characteristics, an improvement in mIoU scores was observed compared to using a single color space.
Notably, when RGB, LAB, and HSV color spaces were combined, the mIoU score improved by approximately 0.03 compared to using the RGB color space alone.
However, it was confirmed in Table 2 that there was an improvement in the mIoU score, while the frequency of undetected events also increased.
This indicates that the choice of color space combinations can significantly impact the detection success rate.
In summary, using appropriate color space combinations can help reduce the rate of undetected underwater glare regions.
However, in some combinations, despite superior qualitative results, lower quantitative scores were observed.
This is because the manually created ground truth regions had ambiguous boundaries, leading to simplified boundary delineation.
As a result of this experiment, it was confirmed that the fusion of image color space channels in pixel clustering-based glare region detection for underwater diver images improved detection results compared to using only RGB images.
In particular, when fusing the Green, Blue, and Lightness channels from the RGB and Lab color spaces, the detection accuracy improved by approximately 0.13 in terms of mIoU compared to RGB images alone.
This combination achieved the highest mIoU score among the tested configurations.
Additionally, in the analysis of the undetection rate for different combinations of image color space fusion, the fusion of Blue and Lightness Value channels resulted in an undetection rate of 8.23 percent at the mIoU threshold of 0.1, which was approximately half of the undetection rate observed with RGB images, making it the most effective combination in reducing undetected cases.

\section{Conclusion}
In this study, we proposed a glare detection algorithm for underwater camera images based on K-means clustering, utilizing image color space fusion and relative coordinate information.
By leveraging machine learning-based K-means clustering, underwater glare regions were grouped without requiring a separate training process.
Furthermore, by applying CLAHE, image resizing, and assigning weights to individual channels, ambiguous boundaries between underwater objects and the background were improved.
By incorporating relative coordinate information, breath bubble regions were distinguished from the bottom surface in underwater images, allowing for effective detection of glare areas associated with breath bubbles.
However, the proposed approach is significantly influenced by domain similarity factors within the underwater camera dataset.
In particular, achieving robust results across varying datasets remains a challenge.
To develop a more robust detection model, future research will focus on constructing datasets through domain adaptation and data augmentation based on the proposed algorithm.


\hfill 
 
\hfill

\appendices
\section*{Acknowledgment}

\ifCLASSOPTIONcaptionsoff
  \newpage
\fi

\end{document}